\documentclass{acm_proc_article-sp}
\usepackage{graphicx}
\usepackage{amssymb}
\usepackage{subfigure}
\usepackage{algorithmic}
\usepackage{algorithm}
\numberwithin{algorithm}{section}  

\graphicspath{{figs/}}

\DeclareGraphicsExtensions{.pdf,.png,.jpg,.mps}

\newcommand{\pair}[2]{<\hspace{-3pt}{#1},{#2}\hspace{-3pt}>}
\newcommand{\xy}{\pair{x}{y}}
\newcommand{\yx}{\pair{y}{x}}

\newtheorem{theorem}{Theorem}[section]

\begin{document}
\title{SortNet: Learning To Rank\\By a Neural-Based Sorting Algorithm}
\numberofauthors{1}
\author{
\alignauthor Leonardo Rigutini, Tiziano Papini, Marco Maggini, Franco Scarselli\\
       \affaddr{Dipartimento di Ingegneria dell'Informazione}\\
       \affaddr{via Roma 56, Siena, Italy}\\
       \email{\{rigutini,papinit,maggini,franco\}@dii.unisi.it}
}
\maketitle
\begin{abstract}
The problem of relevance ranking consists of sorting a set of objects
with respect to a given  criterion. Since users may prefer different relevance criteria, the ranking algorithms should be adaptable to the user needs. Two main approaches exist in literature for the task of learning to rank:
1) a score function, learned by examples, which evaluates the properties of each object yielding an absolute relevance value that can be used to order the objects or 2) a pairwise approach, where a ``preference function'' is learned using pairs of objects to define which one has to be ranked first. In this paper, we present SortNet, an adaptive ranking algorithm which orders objects using a neural network as a comparator. The neural network training  set provides examples of the desired ordering between pairs of items and it is constructed by an iterative procedure which, at each iteration, adds the most informative training examples. 
Moreover, the comparator adopts a connectionist architecture that is particularly suited
for implementing a preference function. We also prove that such an architecture has the universal approximation property and can implement a wide class of  functions.
Finally, the proposed algorithm is evaluated on the LETOR dataset showing promising perfor\-mances in comparison with other state of the art algorithms. 
\end{abstract}

\section{Introduction}
\label{sec:intro}
The standard classification or regression tasks do not include all the supervised
learning problems. Some applications require to focus on other computed
properties of the items, rather than values or classes. For instance, in ranking tasks, the  score value assigned to each object is  less important than the ordering induced on the set of items by the scores. In other cases, the main goal is to retrieve the top $k$ objects without considering the ordering for the remaining items. The differences among these classes of problems influences  the properties of that we would like to predict, 
 the represen\-tation of the patterns and the type of the available supervision. For example, when an user indicates that an object is to be preferred with respect to another, or that two objects should be in the same class, he/she does not assign a value to the objects themselves. In these cases, the given examples are in the form of relationships on pairs of objects and the supervision values are the result of a preference or similarity function applied on the pair of items. 
Two of these peculiar supervised learning tasks are {\em preference learning} and {\em learning to rank}. In the machine learning literature, preference learning problems can be categorized into two specific cases, the {\em Learning Objects Preference} and the {\em Learning Labels Preference}  formulations as reviewed in \cite{PreferenceLearning05}. In the learning objects preferences scenario, it is supposed to have a collection of instances ${x_i}$ with associated a total or partial ordering. The goal of the training is to learn a function that, given a pair of objects, correctly predicts the associated preference as provided by the available ordering. In this approach, the training examples consist of preferences between pairwise instances, while the supervision label consist of the preference expressed by the user on the given pair: $x_i \succ x_j$ if $x_i$ is to be preferred to $x_j$, $x_i \prec x_j$ vice versa. This approach is known as {\em pairwise preference learning} since it is based on pairs of objects. \\
On the other hand, the task of relevance ranking consists  of sorting a set of objects
with respect to a given  criterion. In learning to rank, the criterion is not predefined,
but it has to be adapted to the users' needs. The two research areas of preference learning
and learning to rank have shown many interactions. In particular, the approach of
 Herbrich et al. in \cite{herbrich98learning}, which is based on a binary classifier,  is considered the first work on preference learning and learning to rank. Recently, an increasing number of new algorithms have been proposed to learn a scoring function for ranking objects. Freund et al. \cite{RankBoost98} proposed RankBoost, an algorithm based on a collaborative filtering approach. Burges et al. \cite{RankNet05} used a neural network to model the underlying ranking function (RankNet). Similarly to the approach proposed in this paper, it uses a gradient descent technique to optimize a probabilistic cost function, the cross entropy. The neural network is trained on pairs of training examples using a modified back\-propagation algorithm. It differs from the method proposed in this paper for the weight-sharing scheme and for the training set construction procedure. In \cite{RankSVM06}, the authors use a pairwise learning approach to train a SVM model (SVMRank), while AdaRank \cite{AdaRank07} uses an AdaBoost-based scheme to learn the preference function for ranking. Finally, Zhe Cao et al. proposed ListNet \cite{ListNet07}, that, for the first time, extends the pairwise approach to a listwise approach. In the latter approach, lists of objects are used as instances for learning. 

In this paper we propose SortNet, a ranking algorithm that orders objects
using a neural network as a ``comparator''. The neural network is trained by examples
to learn a compa\-rison function that specifies for each pair of objects which is the preferred one. The network is embedded into a sorting algorithm to provide the ranking of a set of objects. 

The comparator adopts a particular neural architecture that allows us
to implement the symmetries naturally present in a preference function.
The approximation capability of this architecture has been studied
proving that the comparator is an universal approximator and it can implement a
wide class of functions.  The comparator is trained by an iterative procedure,
which aims at selecting the most informative patterns in the training set. 
In particular, at each iteration, the neural network is trained only a subset
of the original training set. This subset is enlarged at each step  by including
the miss-classified patterns. The  procedures selects the comparator that obtains
the best performance on the validation set during the learning procedure.

The proposed approach is evaluated using the LETOR (LEar\-ning TO Rank) dataset \cite{Letor2007}, which is a standard benchmark for the task of learning to rank. The comparison considers several state-of-the-art ranking algorithms, such as RankSVM \cite{RankSVM06}, RankBoost \cite{RankBoost98}, FRank \cite{FRank07}, ListNet \cite{ListNet07}, and AdaRank \cite{AdaRank07}. The paper is organized as follows. In the next section, we introduce the neural network model  along with a brief mathematical description of its properties. In Section \ref{sec:ranking}, the whole SortNet algorithm based on the comparator is defined. In Section \ref{sec:experiments}, we present the experimental setup, the LETOR dataset, and some comparative experimental results. Finally, in Section \ref{sec:conclusions}, some conclusions are drawn.
\section{The neural comparator}
\label{sec:model}

\begin{figure}[h!]
\centering
   \includegraphics[width=7cm]{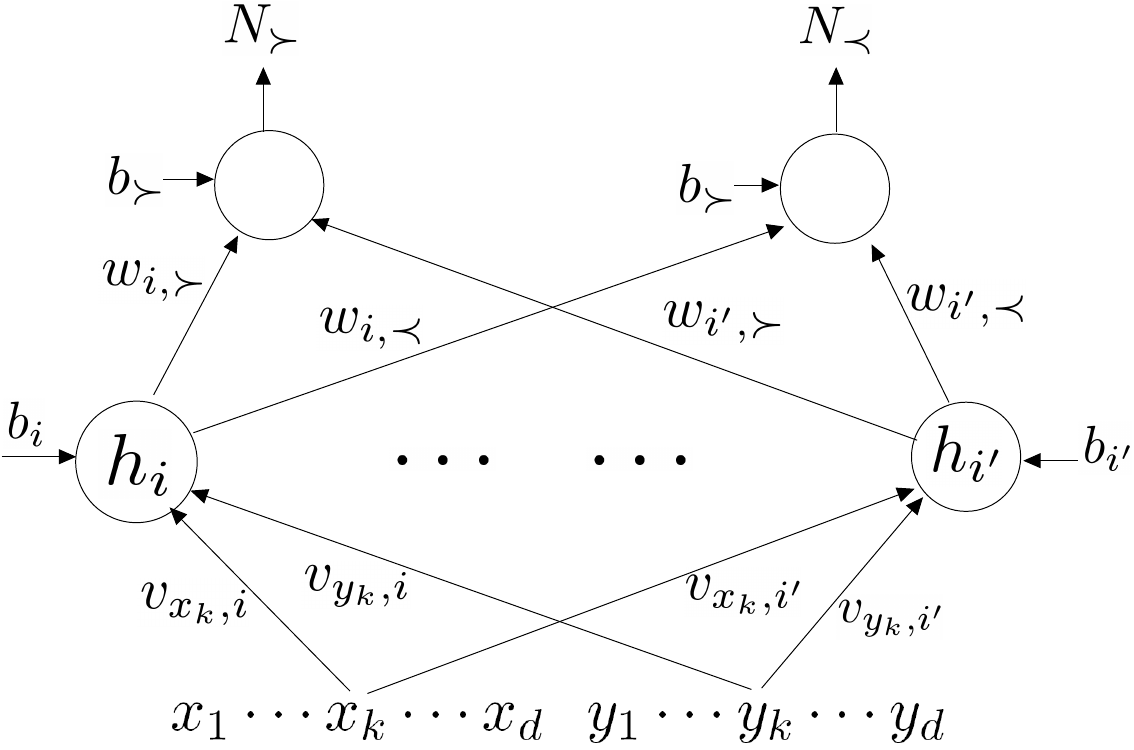}
 \caption{The comparator network architecture}
\label{fig:ann} 
\end{figure}

In the following, $S$ is a set of objects described by a vector of features. We assume that a preference relationship exists between the objects and it is represented by the symbols $\succ$ and $\prec$. Thus, $x \succ y$ means that $x$ is preferred to $y$ while $x \prec y$ that $y$ is preferred to $x$. The purpose of the proposed method is to learn by examples the partial order specified by the preference relationship. The main idea is that of designing a neural network $N$ that processes a representation of two objects $x,y$ and  produces an estimate of $P(x \succ y)$ and $P(x\prec y)$. We will refer to the network as the ``comparator''. More formally, we will have
\begin{eqnarray*}
N_\succ(\xy) & \approx & P(x \succ y)\\  
N_\prec(\xy) & \approx & P(x \prec y),  
\end{eqnarray*}
where  $\xy=[x_1,\ldots,x_d,y_1,\ldots,y_d]$ is the concatenation of the feature vectors of objects $x,y$ and $N_\succ,N_\prec$ denote the two network outputs. Since the neural network approximates a preference function, it is naturally to enforce the following constraints on the outputs:
\begin{eqnarray}
 N_\succ(\xy) =N_\prec(\yx) \ .
\label{eqn:simmetry}
\end{eqnarray}
Equation (\ref{eqn:simmetry}) suggests that the outputs $N_\succ$ and $N_\prec$ must be symmetrical with respect to the order of the examples in the input pair. The comparator consists of a feedforward neural network with one hidden layer, two outputs, implementing $N_\succ$ and $N_\prec$, respectively, and $2d$ input neurons, where $d$ is the dimension of the object feature vectors (see Figure~\ref{fig:ann}). Let us assume that  $v_{x_k,i}$ ($v_{y_k,i}$) denotes the weight of the connection from the input node $x_k$ ($y_k$) \footnote{Here, with an abuse of notation, $x_k$ represents the node that is feeded with  the $k$-th feature of $x$.} to the $i$-th hidden node, $w_{i,\succ}$, $w_{i,\prec}$ represent the weights of the connections from the $i$-th hidden to the output nodes, $b_i$ is the bias of $i$-th hidden and $b_\succ,b_\prec$ are the output biases.
The network architecture adopts a weight sharing mechanism in order to ensure that the constraint (\ref{eqn:simmetry}) holds. For each hidden neuron $i$, a dual neuron $i^{\prime}$ exists whose weights are shared with $i$ according to the following schema:
\begin{enumerate}\label{pagen:const}
\vspace{-0.5 cm}
\item $v_{x_k,i^\prime}=v_{y_k,i}$ and   $v_{y_k,i^\prime}=v_{x_k,i}$ hold, i.e., the weights from $x_k,y_k$ to $i$ are inverted in the connections to  $i^{\prime}$;
\item $w_{i^\prime,\succ}=w_{i,\prec}$ and $w_{i^\prime, \prec}=w_{i,\succ}$ hold, i.e., the weights of the connections from hidden $i$ to outputs $\succ,\prec$ are inverted in the connections leaving from  $i^\prime$;
\item $b_i=b_{i^\prime}$ and $b_\succ=b_\prec$ hold, i.e., the biases are shared between the dual hiddens $i$ and $i^\prime$ and between the outputs $\succ$ and $\prec$.
\end{enumerate}
\vspace{-0.5 cm}
In order to study the properties of the above described architecture, let us denote by $h_i(\xy)$  the output of the $i$-th hidden neuron when the network is feeded on the pair $\xy$. Then, using the weight--sharing  rule in point 1, we have
\begin{eqnarray*}
h_i(\xy)&=&\sigma \left(\sum_k (v_{x_k,i} \,x_k + v_{y_k,i} \,y_k) + b_i \right) \\
&=&\sigma\left(\sum_k (v_{x_k,i^\prime} \, y_k + v_{y_k,i^\prime}\, x_k) + b_i\right)\\
&=&h_{i^\prime}(\yx)
\end{eqnarray*}
where $\sigma$ is the activation function of hidden units. Moreover, let $N_\succ(\xy)$ and $N_\prec(\xy)$ represent the network outputs. Then, by the rules in points 2 and 4,  it follows
\begin{eqnarray*}
&&N_\succ(\xy)=\\
&&=\sigma \left(\sum_{i,i^\prime} (w_{i,\succ}\, h_i(\xy) + w_{i^\prime,\succ} \, h_{i^\prime}(\xy) + b_\succ \right) \\ 
&&=\sigma \left(\sum_{i,i^\prime} (w_{i^\prime,\prec}\, h_{i^\prime}(\yx) + w_{i,\prec} \, h_i(\yx) + b_\prec \right)\\
&& =N_\prec(\yx)\,,
\end{eqnarray*}
where we applied the fact that $h_i(\xy)=h_{i^\prime}(\yx)$ as shown before. Thus,  $N_\succ(\xy)=N_\prec(\yx)$ holds, which proves that the constraint of equation (\ref{eqn:simmetry}) is fulfilled by the network output. 

\subsection*{Approximation capability}
It has been proved that three layered networks are universal approximators~\cite{Hornik89,Funahashi89,Cybenko89}. Similarly, it can be shown that the neural comparator described in this paper can approximate up to any degree of precision most of the practically useful functions that satisfy the constraint (\ref{eqn:simmetry}). Formally, let $\cal F$ be a set of functions $f: R^n \rightarrow R^m$,  $\|\cdot\|$ be a norm on $\cal F$ and $\sigma$ be an activation function. The universal approximation property, proved for three layered networks,  states that for any function $f\in {\cal F}$ and any real $\varepsilon>0$, there exists a network that implements a function  $N: R^n \rightarrow R^m$ such that $\|f-N\|\leq \varepsilon$ holds. Different versions of this property have been proved, according to the adopted function set $\cal F$  (e.g., the set of the continuous or the measurable functions), the norm $\|\cdot\|$ (e.g., the infinity norm or a probability norm) and the activation function $\sigma$ (e.g., a sigmoidal or a threshold activation function)~\cite{ScarselliNN98}. The following theorem demonstrates that the network with the proposed weight sharing mantains the universal approximation property provided that we restrict the attention to the functions that satisfy the constraint (\ref{eqn:simmetry}).
\begin{theorem}\label{theon:app}
Let $\cal F$ be a set of functions $f: R^{2d} \rightarrow R^2$,  $\|\cdot\|$ be a norm on $\cal F$ and $\sigma$ be an activation function such that the corresponding three layered neural network class has the universal approximation property on $\cal F$. Let us denote by  $\overline{\cal F}$ the set of the functions  $k$ that belongs to $\cal F$ and, for any $\xy$, fulfill
\begin{equation}
 k_\succ(\xy) =k_\prec(\yx)
\label{eqn:fconst}
\end{equation}
where $k_\succ,k_\prec$
denote the two components of the outputs of $k$. Then, for any function $f\in \overline{\cal F}$ and any real $\varepsilon>0$, there exists a three layered neural network satisfying the weight sharing schema defined in points 1-4 such that 
$$
\|f-h\|\leq \varepsilon
$$ holds, where  $h: R^{2d} \rightarrow R^2$ is the function implemented by the neural network.
\end{theorem}
\proof (sketch)\\
By the universal approximation hypothesis, there exists a three layered neural network ${\cal A}$ that implements a function $r: R^{2d} \rightarrow R^2$ such $\|\frac{f}{2}-r\|\leq \frac{\varepsilon}{2}$, where $f$ and $\varepsilon$ are the function and the real of the hypothesis, respectively. Then, we can construct another network ${\cal B}$ that has twice the hidden nodes of ${\cal A}$. The indexes of hidden nodes of ${\cal B}$ are  partitioned into pairs $i,i^\prime$. The neurons with index $i$ are connected to the input and to the outputs  with the same weights as in ${\cal A}$: in other words, ${\cal B}$ contains ${\cal A}$ as a sub--network. On the other hand, the weights of the hidden neurons with index $i^\prime$ are defined following the weight sharing rules in points 1,2, and 3. Finally, the biases of the outputs nodes in  ${\cal B}$ are set to twice the values of the biases in ${\cal A}$. 

Notice that, by construction, the network ${\cal B}$  satisfies all the  rules 
of points 1-4 and it is a good candidate to be the network of the thesis. Moreover, 
${\cal B}$ is composed by two sub--networks, the sub-network identified by the hiddens $i$ and the sub--network identified by the hiddens $i^\prime$. Let $p^1, p^2$ represent the functions that define the contribution to the output by the former and the latter sub-networks, respectively. Then, we can easily prove that $p_1$ produces a contribution to output which is equal to $r(\xy)$, i.e., $p^1_\succ(\xy)=r_\succ(\xy)$ and $p^1_\prec(\xy)=r_\prec(\xy)$. On the other hand, $p^2$ has a symmetrical behaviour with respect to $r$ due to the weight--sharing schema, i.e., $p^2_\succ(\xy)=r_\prec(\yx)$ and $p^2_\prec(\xy)=r_\succ(\yx)$. Since, the output function $h$ implemented by ${\cal B}$  is given by the sum of the two components, then
\begin{eqnarray*}
h_\succ(\xy)&=&p^1_\succ(\xy)+p^2_\succ(\xy)= \\
&=&r_\succ(\xy)+r_\prec(\yx)=2r_\succ(\xy)\\ \\
h_\prec(\xy)&=&p^1_\prec(\xy)+p^2_\prec(\xy)= \\
&=&r_\prec(\xy)+r_\succ(\yx)=2r_\prec(\xy)\\ \\
\end{eqnarray*}
where we have used $r_\succ(<x,y>) =r_\prec(<y,x>)$ that holds by definition of $r$ and $f$. Then, the thesis follows straightforwardly by
$$
\|f-h\|= 2\left\|\frac{f}{2}-\frac{h}{2}\right\|=2\left\|\frac{f}{2}-r\right\|\leq\varepsilon
$$
\qed

\subsection*{The training  and the test phase}
To train the comparator, a learning algorithm based on gradient descent is used. For each pair of inputs $\xy$, the assigned target is
\begin{equation}
t = \left\{
\begin{array}{rr}
\left[1 \  0\right] & \mbox{if} \  x \succ y \\
\left[0 \  1\right] & \mbox{if} \  x \prec y
\end{array}
\right. \ .
\label{eq:target}
\end{equation}
and the error is measured by the squared error function
$$
E(\xy)=(t_1-N_\succ(\xy))^2+(t_2-N_\prec(\xy))^2\,.
$$
After training, the comparator can be used to predict the preference relationship between any
pair of objects. Formally, we can define $\succ,\prec$ by
\begin{eqnarray}
 x \succ y &\mbox{  if  } & N_\succ(\xy) > N_\prec(\xy)\nonumber\\
 x \prec y &\mbox{  if  }& N_\prec(\xy) > N_\succ(\xy) \ . \nonumber 
\end{eqnarray}
Notice  that this approach cannot ensure that the predicted relationship defines a total order, since the transitivity property may not hold
\begin{eqnarray*}
\mbox{Transitivity: } x \succ y \mbox{ and } y \succ z \Longrightarrow x \succ z\,.
\end{eqnarray*}
However, the experimental results show that the neural network can easily learn the transitivity relation provided that the training set supports its validity.

\section{The sorting algorithm}
\label{sec:ranking}
The neural comparator is employed to provide a ranking of a set of objects. In particular, the objects are ranked by a common sorting algorithm that exploits the neural network as a comparison function. In this case, the time computational cost of the ranking is mainly due to the sorting algorithm, so that the objects can be ranked in $O(n \log n)$.  It is worth mentioning that the obtained ranking may depend on the initial order of the objects and on the adopted sorting algorithm, since we cannot ensure that the relation $\succ$ is  transitive.  However,  in our experiments, the same sorting algorithm was applied to different shufflings of the same objects and different sorting algorithms were used on the same set of objects. The obtained  orderings were the same in most of the cases and seldom they differed only by a very small number of object positions ($1-5$ over $1000$).
\subsection{The incremental learning procedure}
\label{sec:letor}
Let us assume that a function $RankQuality$ is available to measure
the quality of a given ranking. Some possible choices for this measure
are described in the following subsection. It is worth noticing that
the quality of a given ranked list roughly depends on how many pairs of objects
are correctly ordered. In fact, the comparator neural network is trained
using the square error function $E$, that forces the network
outputs to be close to the desired targets. When
the comparator produces a perfect classification of any input pair,
also the ranking algorithm yields a perfect sorting of the objects.
However, in general, the optimization of the square error does not
necessarily correspond to a good ranking if the example pairs
in the training set are not properly chosen.

In order to optimize the selection of the training pairs with the aim of
improving the ranking performance of the algorithm by using a minimal
number of training examples,  we defined an incremental learning procedure
to train the comparator. This procedure aims at constructing the training set
incrementally by selecting only the pairs of objects that are likely to
be relevant for the goal of obtaining a good ranking. This technique allows
us to optimize the size of the training set avoiding the need to consider
all the possible pairs of the available objects. This method is somehow
related to active learning algorithms, where the learning agent
is allowed to prompt the supervisor for providing the most promising examples.
 
Given the set of objects $T$ and $V$, that can be used as training and validation sets, respectively, at each iteration $i$ a comparator $C^i$ is trained using two subsets $TP \subset T \times T \times \{\prec, \succ\}$, $VP \subset V \times V\times \{\prec, \succ\}$ that contain the current training and validation pairs of the form $x  \succ  y$ or $x  \prec  y$.  In particular $TP$ contains the pairs that are used to train the comparator, while $VP$ is used as validation set for the neural network training procedure. The trained comparator neural network $C^{i+1}$ is used to sort the objects in $T$ and $V$ and the objects pairs that have been mis-compared by the comparator when producing the $R_{T}^i$ are added to the subsets $TP$, $VP$ obtaining the training and validation sets for the next iteration. The procedure is repeated until a maximum number of iterations is reached or until there is no difference in the sets $TP$ and $VP$ between two consecutive iterations. The output of the incremental training is the comparator network $C^*$ that yields the best performance on the validation set during the iterations.
\begin{algorithm}                      
\caption{The SortNet algorithm}          
\label{alg:training}                            
\begin{algorithmic}[1]
	\STATE $T \leftarrow Set\ of\ training\ objects$
	\STATE $V \leftarrow Set\ of\ validation\ objects$
	\STATE $C^0 \leftarrow  randomInit();$ 
	\STATE $TP \leftarrow  \{\};$
	\STATE $VP \leftarrow  \{\};$
	\FOR{$i$ = 0 to max\_iter} 
		\STATE $[TP_i\ ,\ R_{T}^i] \leftarrow Sort(C^i,T);$
		\STATE $[VP_i\ ,\ R_{V}^i] \leftarrow Sort(C^i,V);$
		\STATE $score \leftarrow  RankQuality(R_{V}^i);$
		\IF{$score> best\_score$}
			\STATE $best\_score \leftarrow score;$
			\STATE $C^* \leftarrow C^i;$
		\ENDIF
		\IF{$TP_i\subseteq TP$ and $VP_i\subseteq VP$}
			\STATE return $C^*$;
		\ENDIF
		\STATE $TP \leftarrow  TP \cup TP_i$;
		\STATE $VP \leftarrow  VP \cup VP_i$;
		\STATE $C^{i+1} \leftarrow TrainAndValidate(TP,VP);$
	\ENDFOR 
	\STATE return $C^*$;
\end{algorithmic}
\end{algorithm}

The algorithm is formally described in figure \ref{alg:training}. At the beginning, the neural comparator is randomly initialized and $TP$ and $VP$ are empty. At each iteration $i$, the whole labeled object set $T$ is ranked using the comparator $C^i$ (line $7$). The sorting algorithm returns $R_{T}^i$, that is the ranking of the training examples and $TP_i$, the set of the objects pairs that have been mis-compared by the comparator used in the sorting algorithm to produce $R_T^i$. In detail, the sorting algorithm employs $C^i$ in $|T|\times\log\left(|T|\right)$ pairwise comparisons to produce $R_T^i$, where $|T|$ indicates the size of $T$: the objects pairs for which the $C^i$ output differs from the known correct order are inserted in $TP_i$. The known relative position for the two objects in the pair is available in the set $T$ by exploiting the fact that relevant objects should precede not relevant objects.
Subsequently, in a similar way, $C^i$ is also used to rank the validation set (line $8$), producing a ranking $R_{V}^i$ and a set of misclassified pairs $VP_i$. Then, the set of misclassified pairs $TP_i$ are added to the current learning set $TP$, removing the eventual duplicates (line $14$). Similarly, the pairs in the set $VP_i$ are inserted into $VP$ (line $15$). A new neural network comparator $C^{i+1}$ is trained at each iteration using the current  training and validation sets, $TP$ and  $VP$ (line $16$). More precisely, the neural network training lasts for a predefined number of epochs on the training set $TP$. The selected comparator is the one that achieves the best result, over all the epochs, on the validation set $VP$.
\subsection{Measures of ranking quality}
At each iteration, the quality of the ranking $R_{V}^i$ over the validation set is evaluated by the \emph{RankQuality} function and the model achieving the best performance ($C^*$) is stored (lines $9-13$). In this work,  the following three ranking measures, proposed in the LETOR report \cite{Letor2007}, were used.
\begin{itemize}
\vspace{-0.4cm}
	\item \textbf{Precision at position n (P@n)} --- This value measures the relevance of the top $n$ results of the ranking list with respect to a given query.
$$
P@n=\frac{\textrm{ relevant docs in top}\  n \ \textrm{results}}{n}
$$
\vspace{-0.5cm}
	\item \textbf{Mean average precision (MAP)} --- Given a query $q$, the average precision is 
$$
AP_{q}=\frac{\sum_{n=1}^{N_q} P@n \cdot rel(n)}{\textrm{total relevant docs for} \ q}
$$ 
	where $N_q$ is the number of documents in the result set of $q$ and $rel(n)$ is 1 if the $n$-th document in the ordering is relevant and $0$ otherwise. Thus, $AP_q$ averages the values of  $P@n$ over the positions $n$ of the relevant documents. Finally, the MAP value is computed as the mean of $AP_q$ over the set of all queries.
\vspace{-0.2cm}
\item \textbf{Normalized discount cumulative gain (NDCG@n)} --- This measure exploits an esplicit rating of the documents in the list. The NDCG value of a ranking list at position $n$ is calculated as  
$$
NDCG@n \equiv Z_{n}\sum_{j=1}^n\frac{2^{r_j}-1}{log(1+j)}
$$ 
	where $r_j$ is the rating of the $j$-th document\footnote{Here, it is assumed that $r_j\geq0$ and smaller values indicate less relevance.}, and  $Z_{n}$ is a normalization factor chosen such that the ideal ordering (the DGC-maximizing one) gets a NDCG@n score of 1.
\end{itemize}
\vspace{-0.5cm}
The measure can be choosen according to the user requirements.
For example, if an information retrieval system displays only the best $k$ documents,
 then $P@k$ might be the preferred measure. If, on the other hand, the user is interested on an overall precision of the results, then $MAP$ might be the best choice. Obviously, the selection of a measure to implement \emph{RankQuality} affects the performance of the model on the test set: for example, by selecting the $P@k$ measure, we will obtain a model producing high values of $P@k$ but not optimal values of $MAP$. 
\section{Experimental results}
\label{sec:experiments}
The proposed approach has been validated on two benchmarks, TD2003 and TD2004, which are  included in the LETOR dataset \cite{Letor2007} (LEarning TO Rank)\footnote{The dataset was released by Microsoft Research Asia and is available on line at http://research.microsoft.com/users/LETOR/}. The  benchmarks contain query-document pairs collected from TREC (Text REtrieval Conference). TD2003 consists of $50$ sets of documents, each one containing  $1000$  documents returned in response to a query. Similarly, TD2004 contains $75$ sets of documents corresponding to different queries. The datasets are partitioned into five subsets in order to allow a comparison of different approaches by 5-fold cross-validation: in each experiment, one of the subsets is used for the test, one for validation purposes and three subsets for the training procedure. Each query-document pair is represented by $44$ values which include  several features commonly used in information retrieval. The features also contain a label that specifies whether the document is relevant $R$ or not relevant $NR$ with respect to the query. For all queries, the relevant documents are roughly  $1\%$ of  whole set of documents.

In the reported experiments, the training set was built by considering all the pairs $\xy$ of documents, where $x$ belongs to one of the two relevance classes ($R$ and $NR$) and $y$ belongs to the other class. The corresponding targets  were
\begin{eqnarray*}
t= \left\{
\begin{array}{l}
\mbox{[1 0] if  $x \in R$ and $y \in NR$ }\\
\mbox{[0 1] if  $y \in R$ and $x \in NR$ }\\
\end{array}
\right.
\end{eqnarray*}
The features were normalized to the range $[-1, +1]$ with zero mean. More precisely, let $N_i$ be the number of documents returned in response to the  $i$-th query, and let us denote by $\hat{x}_{j,r}^i$ the normalized $r$-th feature of the $j$-th document  of the $i$-th query, then,
\begin{equation}
\label{norm_features} 
\hat{x}_{j,r}^i=\frac{ x_{j,r}^i-\mu_r^i}{max_{s=1, ..., n_i} |x_{s,r}^i|}\,,
\end{equation}
where  $x_{j,r}^i$ is the original document feature and $\mu_r^i=\frac{\sum_{s=1}^{N_i}x_{s,r}^i}{N_i}$.

In the set of experiments, the documents of the datasets were ranked using the SortNet algorithm. In this setting, a good ranking is characterized by the presence of the relevant documents in the top positions and it was evaluated by the three measures P@n, MAP and NDCG@n. The results obtained by our approach were compared to those achieved by the methods reported in \cite{Letor2007}, i.e., RankSVM \cite{RankSVM06}, RankBoost \cite{RankBoost98}, FRank \cite{FRank07}, ListNet \cite{ListNet07} and AdaRank \cite{AdaRank07}. 

As first trial, we tested the SortNet algorithm varying the number of hidden neurons in $[10,20,30]$ to select the best architecture for the TD2003 and for the TD2004 datasets. In this first set of experiments, the $MAP$ score was used as $RankQuality$ function (line $9$ in algorithm \ref{alg:training}). Table \ref{tbl:architecture_MAP} reports the performances of the three architectures on the test set.
\begin{table*}[htbp]
\centering
\subtable[NDCG@n]
{
\vspace{-1cm}
\hspace{-0.5cm}
\resizebox{16cm}{!}{
\begin{tabular}{|l|c|c|c|c|c|c|c|c|c|c|}
\hline
\textbf{TD2003} & 
\multicolumn{1}{l|}{\textbf{\hspace{0,2cm}n=1\hspace{0,2cm}}} & \multicolumn{1}{l|}{\textbf{\hspace{0,2cm}n=2\hspace{0,2cm}}} & \multicolumn{1}{l|}{\textbf{\hspace{0,2cm}n=3\hspace{0,2cm}}} & \multicolumn{1}{l|}{\textbf{\hspace{0,2cm}n=4\hspace{0,2cm}}} & \multicolumn{1}{l|}{\textbf{\hspace{0,2cm}n=5\hspace{0,2cm}}} & \multicolumn{1}{l|}{\textbf{\hspace{0,2cm}n=6\hspace{0,2cm}}} & \multicolumn{1}{l|}{\textbf{\hspace{0,2cm}n=7\hspace{0,2cm}}} & \multicolumn{1}{l|}{\textbf{\hspace{0,2cm}n=8\hspace{0,2cm}}} & \multicolumn{1}{l|}{\textbf{\hspace{0,2cm}n=9\hspace{0,2cm}}} & \multicolumn{1}{l|}{\textbf{\hspace{0,2cm}n=10\hspace{0,2cm}}} \\ \hline
\textbf{SortNet 10 Hiddens} & \textbf{0,34} & \textbf{0,36} & \textbf{0,35} & \textbf{0,33} & \textbf{0,33} & \textbf{0,33} & \textbf{0,34} & \textbf{0,34} & \textbf{0,33} & \textbf{0,34} \\ \hline
SortNet 20 Hiddens & 0,3 & 0,31 & 0,3 & 0,3 & 0,29 & 0,28 & 0,28 & 0,27 & 0,27 & 0,27 \\ \hline
SortNet 30 Hiddens & 0,36 & 0,29 & 0,29 & 0,28 & 0,27 & 0,27 & 0,28 & 0,27 & 0,27 & 0,28 \\ \hline
 \hline
\textbf{TD2004} & 
\multicolumn{1}{l|}{\textbf{\hspace{0,2cm}n=1\hspace{0,2cm}}} & \multicolumn{1}{l|}{\textbf{\hspace{0,2cm}n=2\hspace{0,2cm}}} & \multicolumn{1}{l|}{\textbf{\hspace{0,2cm}n=3\hspace{0,2cm}}} & \multicolumn{1}{l|}{\textbf{\hspace{0,2cm}n=4\hspace{0,2cm}}} & \multicolumn{1}{l|}{\textbf{\hspace{0,2cm}n=5\hspace{0,2cm}}} & \multicolumn{1}{l|}{\textbf{\hspace{0,2cm}n=6\hspace{0,2cm}}} & \multicolumn{1}{l|}{\textbf{\hspace{0,2cm}n=7\hspace{0,2cm}}} & \multicolumn{1}{l|}{\textbf{\hspace{0,2cm}n=8\hspace{0,2cm}}} & \multicolumn{1}{l|}{\textbf{\hspace{0,2cm}n=9\hspace{0,2cm}}} & \multicolumn{1}{l|}{\textbf{\hspace{0,2cm}n=10\hspace{0,2cm}}} \\ \hline
SortNet 10 Hiddens & 0,47 & 0,49 & 0,47 & 0,47 & 0,47 & 0,47 & 0,48 & 0,48 & 0,49 & 0,49 \\ \hline
\textbf{SortNet 20 Hiddens} & \textbf{0,47} & \textbf{0,54} & \textbf{0,54} & \textbf{0,52} & \textbf{0,52} & \textbf{0,53} & \textbf{0,53} & \textbf{0,54} & \textbf{0,54} & \textbf{0,55} \\ \hline
SortNet 30 Hiddens & 0,47 & 0,5 & 0,5 & 0,49 & 0,48 & 0,49 & 0,49 & 0,5 & 0,5 & 0,5 \\ \hline
\end{tabular}
}
}
\subtable[P@n]
{
\vspace{-1cm}
\hspace{-0.5cm}
\resizebox{16cm}{!}{
\begin{tabular}{|l|c|c|c|c|c|c|c|c|c|c|}
\hline
\textbf{TD2003} & \multicolumn{1}{l|}{\textbf{\hspace{0,2cm}n=1\hspace{0,2cm}}} & \multicolumn{1}{l|}{\textbf{\hspace{0,2cm}n=2\hspace{0,2cm}}} & \multicolumn{1}{l|}{\textbf{\hspace{0,2cm}n=3\hspace{0,2cm}}} & \multicolumn{1}{l|}{\textbf{\hspace{0,2cm}n=4\hspace{0,2cm}}} & \multicolumn{1}{l|}{\textbf{\hspace{0,2cm}n=5\hspace{0,2cm}}} & \multicolumn{1}{l|}{\textbf{\hspace{0,2cm}n=6\hspace{0,2cm}}} & \multicolumn{1}{l|}{\textbf{\hspace{0,2cm}n=7\hspace{0,2cm}}} & \multicolumn{1}{l|}{\textbf{\hspace{0,2cm}n=8\hspace{0,2cm}}} & \multicolumn{1}{l|}{\textbf{\hspace{0,2cm}n=9\hspace{0,2cm}}} & \multicolumn{1}{l|}{\textbf{\hspace{0,2cm}n=10\hspace{0,2cm}}} \\ \hline
\textbf{SortNet 10 Hiddens} & \textbf{0,34} & \textbf{0,34} & \textbf{0,29} & \textbf{0,25} & \textbf{0,24} & \textbf{0,23} & \textbf{0,23} & \textbf{0,21} & \textbf{0,2} & \textbf{0,19} \\ \hline
SortNet 20 Hiddens & 0,3 & 0,3 & 0,27 & 0,25 & 0,22 & 0,2 & 0,19 & 0,17 & 0,16 & 0,15 \\ \hline
SortNet 30 Hiddens & 0,36 & 0,28 & 0,27 & 0,23 & 0,21 & 0,19 & 0,19 & 0,18 & 0,17 & 0,17 \\ \hline
 \hline
\textbf{TD2004} & \multicolumn{1}{l|}{\textbf{\hspace{0,2cm}n=1\hspace{0,2cm}}} & \multicolumn{1}{l|}{\textbf{\hspace{0,2cm}n=2\hspace{0,2cm}}} & \multicolumn{1}{l|}{\textbf{\hspace{0,2cm}n=3\hspace{0,2cm}}} & \multicolumn{1}{l|}{\textbf{\hspace{0,2cm}n=4\hspace{0,2cm}}} & \multicolumn{1}{l|}{\textbf{\hspace{0,2cm}n=5\hspace{0,2cm}}} & \multicolumn{1}{l|}{\textbf{\hspace{0,2cm}n=6\hspace{0,2cm}}} & \multicolumn{1}{l|}{\textbf{\hspace{0,2cm}n=7\hspace{0,2cm}}} & \multicolumn{1}{l|}{\textbf{\hspace{0,2cm}n=8\hspace{0,2cm}}} & \multicolumn{1}{l|}{\textbf{\hspace{0,2cm}n=9\hspace{0,2cm}}} & \multicolumn{1}{l|}{\textbf{\hspace{0,2cm}n=10\hspace{0,2cm}}} \\ \hline
SortNet 10 Hiddens & 0,47 & 0,45 & 0,39 & 0,37 & 0,35 & 0,33 & 0,32 & 0,3 & 0,28 & 0,27 \\ \hline
\textbf{SortNet 20 Hiddens} & \textbf{0,47} & \textbf{0,5} & \textbf{0,46} & \textbf{0,42} & \textbf{0,38} & \textbf{0,38} & \textbf{0,35} & \textbf{0,33} & \textbf{0,32} & \textbf{0,3} \\ \hline
SortNet 30 Hiddens & 0,47 & 0,47 & 0,42 & 0,38 & 0,35 & 0,34 & 0,32 & 0,31 & 0,28 & 0,26 \\ \hline
\end{tabular}
}
}
\subtable[MAP]
{
\vspace{-1cm}
\hspace{-0.5cm}
\resizebox{5cm}{!}{
\begin{tabular}{|l|c|}
\hline
\textbf{TD2003} & \textbf{MAP} \\ \hline
\textbf{SortNet 10 Hiddens} & \textbf{0,23} \\ \hline
SortNet 20 Hiddens & 0,2 \\ \hline
SortNet 30 Hiddens & 0,21 \\ \hline
\hline
\textbf{TD2004} & \textbf{MAP} \\ \hline
SortNet 10 Hiddens & 0,41 \\ \hline
\textbf{SortNet 20 Hiddens} & \textbf{0,45} \\ \hline
SortNet 30 Hiddens & 0,41 \\ \hline
\end{tabular}
}
}
\caption{The results achieved on TREC2003 and TD2004 varying the hidden neuron number: (a) NDCG@n, (b) P@n and (c) MAP. The algorithm uses $MAP$ as $RankQuality$ function.}
\label{tbl:architecture_MAP}
\end{table*}
The results on the validation set reflect the performances reported in Table \ref{tbl:architecture_MAP} on the test set. Thus, for the subsequent experiments, we selected a 10-hidden comparator for the TD2003 dataset and a 20-hidden comparator for the TD2004 dataset. 

\begin{figure}[h!]
\centering
\subfigure[TREC2003]{
	\includegraphics[width=5cm]{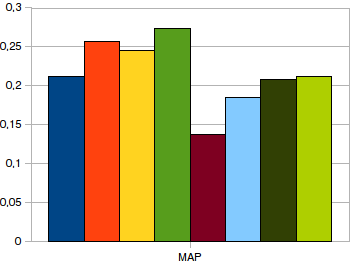}
	\includegraphics[width=2.6cm]{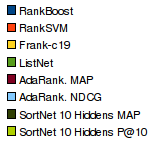}
}
\hspace{0.2mm}
\subfigure[TREC2004]{
	\includegraphics[width=5cm]{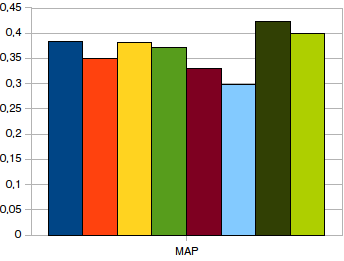}
	\includegraphics[width=2.6cm]{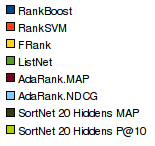}
}
\caption{\footnotesize{MAP on TREC2003 and TD2004.}}
\label{fig:MAP} 
\end{figure}

\begin{figure}[h!]
   \includegraphics[width=8.5cm]{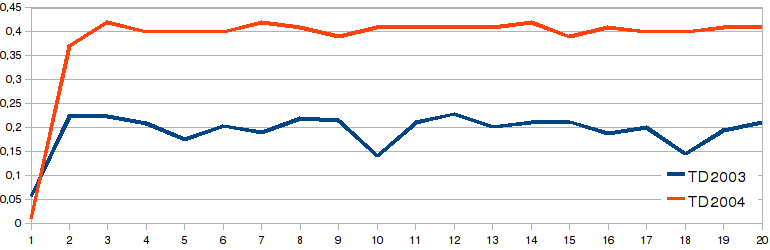}
 \caption{\footnotesize{Trend of MAP on the validation set during the training process for the TREC2003 and TD2004 datasets.}}
\label{fig:MAP_trend}
\end{figure}

After the selection of the best neural network architecture of the comparator, we performed a set of experiments to compare the performances of SortNet with the methods reported in \cite{Letor2007}. In particular, we ran the algorithm on the TD2003 and TD2004 datasets setting $max\_iter=20$ and using $MAP$ and $P@10$ as $RankQuality$ function. 
\begin{table*}[htbp]
\centering
\subtable[TREC2003]
{
\vspace{-1cm}
\hspace{-0.5cm}
\resizebox{16cm}{!}{
\begin{tabular}{|l|c|c|c|c|c|c|c|c|c|c|}
\hline
\textbf{NDCG} & 
\multicolumn{1}{l|}{\textbf{\hspace{0,2cm}n=1\hspace{0,2cm}}} & \multicolumn{1}{l|}{\textbf{\hspace{0,2cm}n=2\hspace{0,2cm}}} & \multicolumn{1}{l|}{\textbf{\hspace{0,2cm}n=3\hspace{0,2cm}}} & \multicolumn{1}{l|}{\textbf{\hspace{0,2cm}n=4\hspace{0,2cm}}} & \multicolumn{1}{l|}{\textbf{\hspace{0,2cm}n=5\hspace{0,2cm}}} & \multicolumn{1}{l|}{\textbf{\hspace{0,2cm}n=6\hspace{0,2cm}}} & \multicolumn{1}{l|}{\textbf{\hspace{0,2cm}n=7\hspace{0,2cm}}} & \multicolumn{1}{l|}{\textbf{\hspace{0,2cm}n=8\hspace{0,2cm}}} & \multicolumn{1}{l|}{\textbf{\hspace{0,2cm}n=9\hspace{0,2cm}}} & \multicolumn{1}{l|}{\textbf{\hspace{0,2cm}n=10\hspace{0,2cm}}} \\ \hline
RankBoost & 0,26 & 0,28 & 0,27 & 0,27 & 0,28 & 0,28 & 0,29 & 0,28 & 0,28 & 0,29 \\ \hline
RankSVM & 0,42 & 0,37 & 0,38 & 0,36 & 0,35 & 0,34 & 0,34 & 0,34 & 0,34 & 0,34 \\ \hline
Frank-c19.0 & 0,44 & 0,39 & 0,37 & 0,34 & 0,33 & 0,33 & 0,33 & 0,33 & 0,34 & 0,34 \\ \hline
ListNet & 0,46 & 0,43 & 0,41 & 0,39 & 0,38 & 0,39 & 0,38 & 0,37 & 0,38 & 0,37 \\ \hline
AdaRank. MAP & 0,42 & 0,32 & 0,29 & 0,27 & 0,24 & 0,23 & 0,22 & 0,21 & 0,2 & 0,19 \\ \hline
AdaRank. NDCG & 0,52 & 0,41 & 0,37 & 0,35 & 0,33 & 0,31 & 0,3 & 0,29 & 0,28 & 0,27 \\ \hline
SortNet 10 Hiddens MAP & 0,38 & 0,3 & 0,28 & 0,29 & 0,29 & 0,3 & 0,29 & 0,29 & 0,29 & 0,28 \\ \hline
SortNet 10 Hiddens P@10 & 0,32 & 0,32 & 0,31 & 0,31 & 0,3 & 0,3 & 0,29 & 0,3 & 0,31 & 0,31 \\ \hline
\hline
\textbf{P@n} & 
\multicolumn{1}{l|}{\textbf{\hspace{0,2cm}n=1\hspace{0,2cm}}} & \multicolumn{1}{l|}{\textbf{\hspace{0,2cm}n=2\hspace{0,2cm}}} & \multicolumn{1}{l|}{\textbf{\hspace{0,2cm}n=3\hspace{0,2cm}}} & \multicolumn{1}{l|}{\textbf{\hspace{0,2cm}n=4\hspace{0,2cm}}} & \multicolumn{1}{l|}{\textbf{\hspace{0,2cm}n=5\hspace{0,2cm}}} & \multicolumn{1}{l|}{\textbf{\hspace{0,2cm}n=6\hspace{0,2cm}}} & \multicolumn{1}{l|}{\textbf{\hspace{0,2cm}n=7\hspace{0,2cm}}} & \multicolumn{1}{l|}{\textbf{\hspace{0,2cm}n=8\hspace{0,2cm}}} & \multicolumn{1}{l|}{\textbf{\hspace{0,2cm}n=9\hspace{0,2cm}}} & \multicolumn{1}{l|}{\textbf{\hspace{0,2cm}n=10\hspace{0,2cm}}} \\ \hline
RankBoost & 0,26 & 0,27 & 0,24 & 0,23 & 0,22 & 0,21 & 0,21 & 0,19 & 0,18 & 0,18 \\ \hline
RankSVM & 0,42 & 0,35 & 0,34 & 0,3 & 0,26 & 0,24 & 0,23 & 0,23 & 0,22 & 0,21 \\ \hline
Frank-c19 & 0,44 & 0,37 & 0,32 & 0,26 & 0,23 & 0,22 & 0,21 & 0,21 & 0,2 & 0,19 \\ \hline
ListNet & 0,46 & 0,42 & 0,36 & 0,31 & 0,29 & 0,28 & 0,26 & 0,24 & 0,23 & 0,22 \\ \hline
AdaRank. MAP & 0,42 & 0,31 & 0,27 & 0,23 & 0,19 & 0,16 & 0,14 & 0,13 & 0,11 & 0,1 \\ \hline
AdaRank. NDCG & 0,52 & 0,4 & 0,35 & 0,31 & 0,27 & 0,24 & 0,21 & 0,19 & 0,17 & 0,16 \\ \hline
SortNet 10 Hiddens MAP & 0,38 & 0,29 & 0,25 & 0,25 & 0,24 & 0,24 & 0,21 & 0,2 & 0,18 & 0,17 \\ \hline
SortNet 10 Hiddens P@10 & 0,32 & 0,31 & 0,29 & 0,27 & 0,25 & 0,23 & 0,21 & 0,21 & 0,2 & 0,2 \\ \hline
\end{tabular}
}
}
\subtable[TREC2004]
{
\vspace{-1cm}
\hspace{-0.5cm}
\resizebox{16cm}{!}{
\begin{tabular}{|l|c|c|c|c|c|c|c|c|c|c|}
\hline
\textbf{NDCG} & \multicolumn{1}{l|}{\textbf{\hspace{0,2cm}n=1\hspace{0,2cm}}} & \multicolumn{1}{l|}{\textbf{\hspace{0,2cm}n=2\hspace{0,2cm}}} & \multicolumn{1}{l|}{\textbf{\hspace{0,2cm}n=3\hspace{0,2cm}}} & \multicolumn{1}{l|}{\textbf{\hspace{0,2cm}n=4\hspace{0,2cm}}} & \multicolumn{1}{l|}{\textbf{\hspace{0,2cm}n=5\hspace{0,2cm}}} & \multicolumn{1}{l|}{\textbf{\hspace{0,2cm}n=6\hspace{0,2cm}}} & \multicolumn{1}{l|}{\textbf{\hspace{0,2cm}n=7\hspace{0,2cm}}} & \multicolumn{1}{l|}{\textbf{\hspace{0,2cm}n=8\hspace{0,2cm}}} & \multicolumn{1}{l|}{\textbf{\hspace{0,2cm}n=9\hspace{0,2cm}}} & \multicolumn{1}{l|}{\textbf{\hspace{0,2cm}n=10\hspace{0,2cm}}} \\ \hline
RankBoost & 0,48 & 0,47 & 0,46 & 0,44 & 0,44 & 0,45 & 0,46 & 0,46 & 0,46 & 0,47 \\ \hline
RankSVM & 0,44 & 0,43 & 0,41 & 0,41 & 0,39 & 0,4 & 0,41 & 0,41 & 0,41 & 0,42 \\ \hline
FRank & 0,44 & 0,47 & 0,45 & 0,43 & 0,44 & 0,45 & 0,46 & 0,45 & 0,46 & 0,47 \\ \hline
ListNet & 0,44 & 0,43 & 0,44 & 0,42 & 0,42 & 0,42 & 0,43 & 0,45 & 0,46 & 0,46 \\ \hline
AdaRank.MAP & 0,41 & 0,39 & 0,4 & 0,39 & 0,39 & 0,4 & 0,4 & 0,4 & 0,4 & 0,41 \\ \hline
AdaRank.NDCG & 0,36 & 0,36 & 0,38 & 0,38 & 0,38 & 0,38 & 0,38 & 0,38 & 0,39 & 0,39 \\ \hline
\textbf{SortNet 20 Hiddens MAP} & \textbf{0,48} & \textbf{0,52} & \textbf{0,53} & \textbf{0,5} & \textbf{0,5} & \textbf{0,5} & \textbf{0,49} & \textbf{0,5} & \textbf{0,51} & \textbf{0,51} \\ \hline
SortNet 20 Hiddens P@10 & 0,43 & 0,5 & 0,47 & 0,47 & 0,46 & 0,47 & 0,47 & 0,48 & 0,48 & 0,49 \\ \hline
\hline
\textbf{P@n} & \multicolumn{1}{l|}{\textbf{\hspace{0,2cm}n=1\hspace{0,2cm}}} & \multicolumn{1}{l|}{\textbf{\hspace{0,2cm}n=2\hspace{0,2cm}}} & \multicolumn{1}{l|}{\textbf{\hspace{0,2cm}n=3\hspace{0,2cm}}} & \multicolumn{1}{l|}{\textbf{\hspace{0,2cm}n=4\hspace{0,2cm}}} & \multicolumn{1}{l|}{\textbf{\hspace{0,2cm}n=5\hspace{0,2cm}}} & \multicolumn{1}{l|}{\textbf{\hspace{0,2cm}n=6\hspace{0,2cm}}} & \multicolumn{1}{l|}{\textbf{\hspace{0,2cm}n=7\hspace{0,2cm}}} & \multicolumn{1}{l|}{\textbf{\hspace{0,2cm}n=8\hspace{0,2cm}}} & \multicolumn{1}{l|}{\textbf{\hspace{0,2cm}n=9\hspace{0,2cm}}} & \multicolumn{1}{l|}{\textbf{\hspace{0,2cm}n=10\hspace{0,2cm}}} \\ \hline
RankBoost & 0,48 & 0,45 & 0,4 & 0,35 & 0,32 & 0,3 & 0,29 & 0,28 & 0,26 & 0,25 \\ \hline
RankSVM & 0,44 & 0,41 & 0,35 & 0,33 & 0,29 & 0,27 & 0,26 & 0,25 & 0,24 & 0,23 \\ \hline
FRank & 0,44 & 0,43 & 0,39 & 0,34 & 0,32 & 0,31 & 0,3 & 0,27 & 0,26 & 0,26 \\ \hline
ListNet & 0,44 & 0,41 & 0,4 & 0,36 & 0,33 & 0,31 & 0,3 & 0,29 & 0,28 & 0,26 \\ \hline
AdaRank.MAP & 0,41 & 0,35 & 0,34 & 0,3 & 0,29 & 0,28 & 0,26 & 0,24 & 0,23 & 0,22 \\ \hline
AdaRank.NDCG & 0,36 & 0,32 & 0,33 & 0,3 & 0,28 & 0,26 & 0,24 & 0,23 & 0,22 & 0,21 \\ \hline
\textbf{SortNet 20 Hiddens MAP} & \textbf{0,48} & \textbf{0,49} & \textbf{0,46} & \textbf{0,4} & \textbf{0,36} & \textbf{0,34} & \textbf{0,3} & \textbf{0,29} & \textbf{0,28} & \textbf{0,27} \\ \hline
SortNet 20 Hiddens P@10 & 0,43 & 0,46 & 0,39 & 0,37 & 0,34 & 0,32 & 0,29 & 0,29 & 0,28 & 0,27 \\ \hline
\end{tabular}
}
}
\caption{\footnotesize{The results achieved on (a) TREC2003 and (b) TD2004.}}
\label{tbl:results}
\end{table*}
Figure \ref{fig:MAP} and the Tables \ref{tbl:results} (a-b) show that, on the TD2004 dataset, the SortNet algorithm clearly outperforms all the other methods. On TD2003, the SortNet method reports values of $MAP$ and $P@n$ similar to the results of the AdaRank and RankBoost. In the figure \ref{fig:MAP_trend}, we report the plot of the $MAP$ value on the validation set at each training iteration: this plot clearly shows that convergence is reached before the maximum number of iterations. 

During the experiments on the TD2003 dataset, however, we noticed a particular behaviour of the algorithm: at each iteration, the performances on the validation set sligthly differed from the performances on the training set and test set. In particular, when the algorithm reported high values for $MAP$ and $P@n$ for the validation set, the $MAP$ and $P@n$ on the test set and the training set were low, and viceversa. This could mean that the distribution of data in the validation clearly differs from the test set and from the training set. This observation is also supported by the fact that in the TD2003 experiments, the neural architecture reporting the best performances has a smaller number of neurons than the best one in the TD2004.

\section{Conclusions}
\label{sec:conclusions}

In this paper a neural-based learning-to-rank algorithm has been proposed. A neural network is trained by examples to decide which of two objects is preferable. The learning set is selected by an iterative procedure which aims to maximize the quality of the ranking. The network adopts a weight sharing schema to ensure that the outputs satisfy logical symmetries which are desirable in an ordering relationship. Moreover, we proved that such a connectionist architecture is a universal approximator.

In order to evaluate the performances of the proposed algorithm, experiments were performed using the datasets TD2003 and TD2004. The results show that our approach outperforms the current state of the art methods on TD2004 and is comparable to  other techniques on
TD2003. It is also argued that a difference on the distribution of the patterns between the validation
and the training and the test set of TD2003 may be responsible for some limitations on the performance of our and other algorithms. 

Matters of future research include a wider experimentation of the approach and the study of different learning procedures  to improve the performance of the method.
\bibliographystyle{abbrv}
\bibliography{letor2008}  
%
\end{document}